\documentclass[a4paper,11pt]{article}

\usepackage{array}
\usepackage[table,xcdraw]{xcolor}

\usepackage{fourier}
\usepackage{color}
 \usepackage{graphicx}
\usepackage{url}
\usepackage[affil-it]{authblk}
\usepackage{amsmath}
\usepackage{wrapfig}

\usepackage[T1]{fontenc}
\usepackage{times}

\begin{document}

\title{Spiking-DD: Neuromorphic Event Camera based Driver Distraction Detection with Spiking Neural Network}
\author[*,1,2]{Waseem Shariff} 
\author[1]{Paul Kielty} 
\author[2]{Joseph Lemley} 
\author[1]{Peter Corcoran} 
\affil[1]{C3I Imaging Lab, Dept. of Electrical and Electronic Engineering, University of Galway, Ireland}
\affil[2]{FotoNation Ltd. Galway, Ireland}
\date{}
\maketitle
\thispagestyle{empty}

\begin{abstract}
Event camera-based driver monitoring is emerging as a pivotal area of research, driven by its significant advantages such as rapid response, low latency, power efficiency, enhanced privacy, and prevention of under-sampling. Effective detection of driver distraction is crucial in driver monitoring systems to enhance road safety and reduce accident rates. The integration of an optimized sensor such as Event Camera with an optimized network is essential for maximizing these benefits. This paper introduces the innovative concept of 'sensing without seeing' to detect driver distraction, leveraging computationally efficient spiking neural networks (SNN). To the best of our knowledge, this study is the first to utilize event camera data with spiking neural networks for driver distraction. The proposed Spiking-DD network not only achieve state-of-the-art performance but also exhibit fewer parameters and provides greater accuracy than current event-based methodologies.

\end{abstract}

\section{Introduction}

Regulatory authorities are increasingly recognizing the critical role of advanced driver monitoring systems in ensuring road safety. For instance, the European Union has issued a directive that, by 2025, all newly produced vehicles must be fitted with driver monitoring systems (DMS) to improve road safety \cite{EuropeanCommission2020}. This regulatory movement underscores the pressing need for innovative and effective solutions to combat driver distraction and enhance overall vehicle safety.

Within the rapidly advancing domain of autonomous and semi-autonomous vehicles, maintaining driver safety is of utmost importance \cite{EuropeanCommission2020}. A key aspect of safety is the ability to accurately detect and address driver distraction, which is a leading contributor to vehicular accidents. Conventional methods of driver monitoring typically depend on continuous video streams and sophisticated machine learning algorithms, which can be both computationally intensive and less efficient in real-time applications \cite{gallego2020event}. Event cameras, unlike conventional frame-based cameras, operate asynchronously and produce data only when changes occur in the visual field. This results in a stream of events that correspond to changes in light intensity at specific points in the field of view. Event cameras do not capture entire images at regular intervals; instead, they detect and record changes in the scene, significantly reducing data redundancy. This method aligns with privacy by design principles and sensing without seeing, as continuous scene reconstruction is unnecessary, minimizing the storage and transmission of potentially sensitive visual data.

When working with optimized data, such as event camera data, it’s essential to pair it with optimized networks for better performance. The quality of both the input data and the neural network architecture significantly impacts the overall results. Spiking Neural Networks (SNNs) \cite{yamazaki2022spiking} represent a different paradigm in neural network design, inspired by the way biological neurons communicate through discrete spikes. Unlike traditional artificial neural networks, which process information in a continuous manner, SNNs operate in an event-driven mode, making them highly efficient for temporal data processing. Leveraging event camera data and the functioning of Spiking Neural Networks (SNNs) underscores the potential of spiking event-based driver distraction detection as a viable alternative \cite{shariff2024event}. The motivation for employing spiking event-based systems in driver distraction detection is multifaceted. The dynamic nature of driving environments necessitates a system capable of rapid and precise responses to sudden changes. Traditional frame-based video analysis can suffer from latency and high computational overhead, making real-time detection challenging. In contrast, spiking event-based systems,  offer a more efficient and responsive solution.

\begin{figure}
    \centering
    \includegraphics[width=1.0\linewidth]{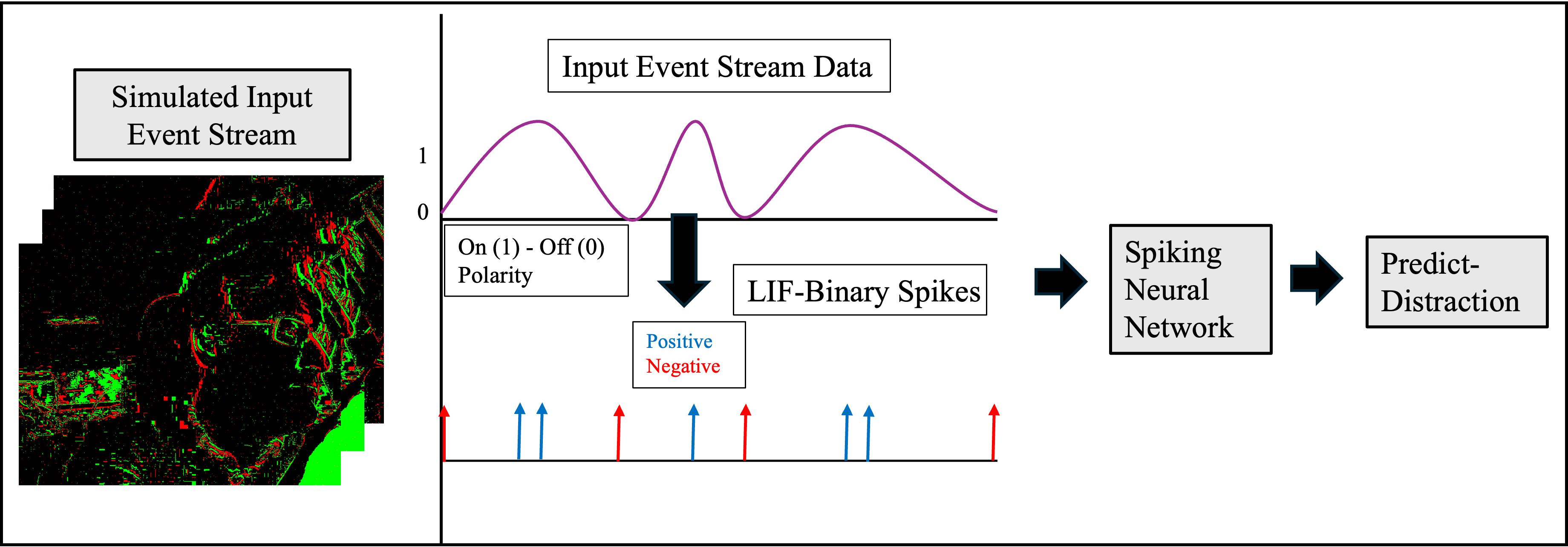}
    \caption{Overview of the proposed sensing pipeline}
    \label{fig:enter-label}
\end{figure}

In this research, as mentioned in figure \ref{fig:enter-label} demonstrates a pipeline where a simulated input event stream is processed and fed into a spiking neural network, which then predicts distractions based on the temporal patterns within the input data. The process begins with a simulated input event stream, which consists of discrete events with on-off polarity. In the LIF neuron model, these events are used to generate binary spikes. Specifically, an ‘on’ event is represented as a spike (binary 1), while an ‘off’ event or a lack of input results in no spike (binary 0). The transformed input stream, now represented as binary spikes, is then fed into the SNN. The network processes these temporal spike patterns to identify features or patterns over time. Finally, the output of the SNN is used to predict distractions, indicating that the network is trained to detect such events.

By focusing on changes rather than static images, spiking event-based systems can provide rapid responses to driver distractions. The asynchronous nature of event cameras means they have negligible latency, which is crucial for real-time applications. Furthermore, the reduced data volume lowers computational requirements, making these systems more suitable for implementation in vehicles where processing power and energy efficiency are critical considerations. Additionally, the privacy advantages of event cameras cannot be overstated. Since these cameras do not capture full images, the risk of misuse of visual data is greatly reduced. This characteristic is particularly important in the context of driver monitoring, where continuous video recording could raise significant privacy concerns.

Contributions:
\begin{itemize}

   \item This work pioneers the use of SNNs, inspired by the efficient and rapid processing capabilities, to detect driver distractions. By leveraging the event-driven nature of SNNs, the system can respond more swiftly and accurately to dynamic changes in the driving environment compared to traditional methods.

    \item This approach not only enhances real-time processing efficiency but also aligns with privacy by design principles, minimizing the need to store and transmit potentially sensitive visual data.

    \item  The network is trained using simulated event streams to test and validate the effectiveness of the SNN-based driver distraction detection system.

\end{itemize}

\section{Background}
Driver distraction has long been a prominent area of research aimed at improving road safety. Traditional approaches to detecting driver distraction have predominantly depended on conventional camera systems. While these methods \cite{canas2021detection}, \cite{lakhani2022applying} have demonstrated a degree of accuracy, they frequently require considerable computational resources, which poses significant challenges for real-time implementation and long-term sustainability. The substantial computational demands arise from the need for continuous frame-by-frame processing inherent in standard camera systems, a process that is both time-consuming and resource-intensive. Consequently, there is an increasing interest in exploring alternative methods that reduce computational demands while maintaining or improving the accuracy of driver distraction detection. Recent advancements in technologies such as event-based cameras and neuromorphic computing offer promising avenues for addressing these challenges and advancing the field of driver distraction detection.

In contrast, event cameras have emerged as a promising alternative for real-time driver distraction detection. Despite the advantages of event cameras, research in this area remains relatively sparse, with only a few studies exploring their potential for distraction detection. One notable study by Yang et al. \cite{Yang} investigated the use of Long Short-Term Memory (LSTM) networks for detecting driver distraction. LSTM networks are a type of recurrent neural network (RNN) capable of learning and remembering long-term dependencies, making them well-suited for time-series data like event camera outputs. In their study, Yang et al. employed the Drive\&Act \cite{driveact} dataset, a comprehensive collection of driver behaviors captured in various scenarios. The LSTM algorithm developed by Yang et al. \cite{Yang} achieved an impressive accuracy of 89\% in detecting driver distraction, demonstrating the potential of event cameras coupled with advanced neural network algorithms.

Another significant contribution to this field comes from Shariff et al. \cite{shariff2023neuromorphic}, who proposed the use of a submanifold sparse convolutional network, specifically the Sparse-ResNet architecture, for driver distraction detection. Sparse-ResNet is designed to handle sparse data efficiently, making it a suitable choice for processing the sparse event data generated by event cameras. Shariff et al. evaluated their model using two different datasets: Drive\& Act \cite{driveact} and the Driver Monitoring Dataset (DMD) \cite{ortega2020dmd}. Their results showed an accuracy of 86.25\% on the Drive\&Act dataset and 80.05\% on the DMD, highlighting the effectiveness of their approach across different datasets and driving conditions.

\section{Methodology}
In this study, we initially simulated DMD video data into event streams using V2E. We then utilized the Leaky-Integrate-and-Fire (LIF) model to generate binary spikes, which were subsequently used to train a spiking neural network with low parameters. The network’s purpose was to classify whether the driver is distracted or not.

\subsection{Event Representation: Leaky Integrate-and-Fire (LIF) Dynamics}

The Leaky Integrate-and-Fire (LIF) model is a fundamental spiking neuron model used in computational neuroscience. It integrates incoming spikes over time and generates an output spike when the membrane potential surpasses a certain threshold. The dynamics of the LIF model are characterized by three primary equations:

Membrane Potential Update:
\begin{equation}
    y[t] = (1 - \alpha) \, y[t-1] + x[t]
\end{equation}
where \( y[t] \) is the current membrane potential, \( y[t-1] \) is the previous membrane potential, \( \alpha \) is the leak factor, and \( x[t] \) is the input at time \( t \). The term \( (1 - \alpha) \) determines the rate at which the previous state decays.

Spike Generation:
\begin{equation}
    s[t] = y[t] \geq \vartheta
\end{equation}
Here, \( s[t] \) is a binary variable indicating whether the current state \( y[t] \) exceeds a threshold \( \vartheta \). If \( y[t] \) is greater than or equal to \( \vartheta \), \( s[t] \) becomes 1 (true); otherwise, it is 0 (false).

Reset Mechanism:
\begin{equation}
    y[t] = y[t] \cdot (1 - s[t])
\end{equation}
This equation ensures that if a spike occurs (\( s[t] = 1 \)), the current state \( y[t] \) is reset to 0. Otherwise, \( y[t] \) remains unchanged.

\subsection{Spiking Neural Network (SNN)}

The Spiking Neural Network (SNN) architecture designed for processing input spikes through multiple layers leverages the LIF neuron model. The network typically includes an input layer, hidden layers, and an output layer. The input layer converts raw data into spike trains using encoding methods such as rate coding or temporal coding. Rate coding represents input values by the frequency of spikes, while temporal coding uses the precise timing of spikes.

\begin{figure}
    \centering
    \includegraphics[width=1.0\linewidth]{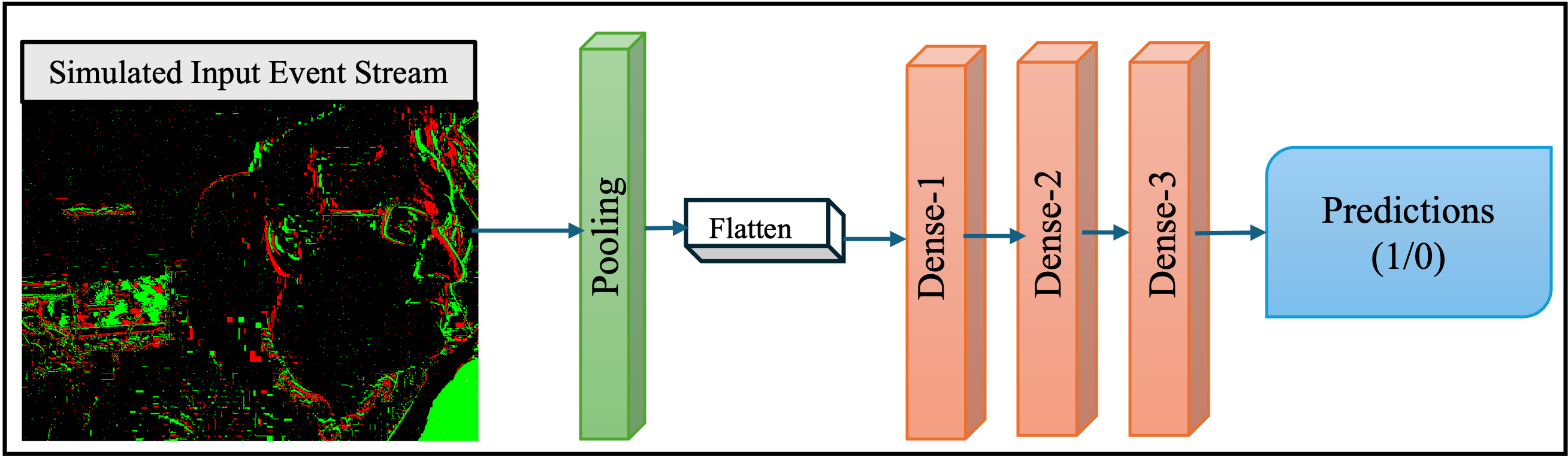}
    \caption{Proposed spiking-DD neural network architecture.}
    \label{fig:network}
\end{figure}

The hidden layers consist of multiple layers of LIF neurons that process the spike trains. Figure \ref{fig:network} shows the network architecture of the proposed spiking neural network. The architecture includes a pooling layer for spatial downsampling, reducing the spatial dimensions of the input without losing a lot of information, followed by a flattening layer to convert the spatially reduced data into a 1D vector. 3X Dense (fully connected) layers further process the spike trains. These layers are parameterized with values such as threshold 1.25, current decay 0.25, voltage decay 0.03, tau grad 0.03, scale grad 3, and dropout 0.05 for regularization. Specifically, the architecture features a pooling layer with a kernel size of 7, a flattening layer, and three dense layers with configurations: 2x69x91 input units to 32 output units, 32 input units to 8 output units, and 8 input units to 2 output units (binary output 0 or 1 for classifying distraction).

During forward propagation, input spikes are fed into the network, and each layer updates its neurons' membrane potentials based on the incoming spikes and their previous states, as described by the LIF dynamics. The spike trains propagate through the network's layers, with each layer's output serving as the input for the next layer, culminating in the final output layer's spike trains.

\subsection{SpikeRate Loss Mechanism}

Training the SNN involves optimizing its performance using the SpikeRate loss mechanism \cite{lava_dl}. This loss function calculates the discrepancy between the actual and target spike rates. The target spike rate \(\hat{\boldsymbol{r}}\) is defined as \( r_{\text{true}} \cdot \mathbf{1}[\text{label}] + r_{\text{false}} \cdot (1 - \mathbf{1}[\text{label}]) \), where \( r_{\text{true}} \) and \( r_{\text{false}} \) represent the true and false spiking rates, respectively.

The loss \(L\) is calculated as:
\begin{equation}
    L = 
    \begin{cases}
        \frac{1}{2} \int_T (\boldsymbol{r}(t) - \hat{\boldsymbol{r}}(t))^\top \mathbf{1} \, \text{d}t & \text{if moving window} \\
        \frac{1}{2} (\boldsymbol{r} - \hat{\boldsymbol{r}})^\top \mathbf{1} & \text{otherwise}
    \end{cases}
\end{equation}

The moving window mechanism allows the loss calculation to consider the temporal aspect of spike trains, enhancing the model's ability to handle temporal data effectively.

In the proposed application of driver distraction detection, 33-millisecond segments of sensor data monitoring driver behavior are collected and converted into spike trains. The loss value in first epoch started at 0.047 with 51\% accuracy and ended at 30 epochs at 0.01074 with 94\% accuracy. These spike trains are processed by the SNN with LIF neurons, which classify each segment as indicating distraction or not based on the spiking patterns in the output layer. The network is trained with collected data, and synaptic weights are adjusted to improve classification accuracy. The LIF model's ability to mimic the temporal dynamics of biological neurons enables SNNs to efficiently process temporal data and make rapid classifications, crucial for real-time applications like driver distraction detection \cite{gallego2020event}, \cite{shariff2024event}.

\section{Experimentation}
The following section will delve into the dataset, explore the simulation of RGB to events, and provide implementation details.

\subsection{Dataset}
In this research, we utilized the Driver Monitoring Dataset (face-focused) \cite{ortega2020dmd}. This dataset is notable for its inclusion of data collected under three distinct scenarios: a laboratory environment, vehicle scenarios, and real-world driving scenarios. The laboratory environment provided a controlled setting for precise experimentation. In contrast, the vehicle scenarios involved capturing data while subjects performed specific activities inside a vehicle. The real-world driving scenarios included data collected during actual driving situations.

The dataset, provided by the authors of \cite{ortega2020dmd}, includes information from 20 subjects, each participating in three different driving scenarios. We utilized the same dataset prepared by \cite{shariff2023neuromorphic}. The dataset consists of 1,680 video samples, each with an average duration of 3 seconds with resolution of 640x480. These rgb videos were simulated to event stream data using the v2e event data simulator \cite{hu2021v2e}. The dataset features a variety of driving scenarios, including distracted driving activities such as using mobile phones, drinking water, fixing hair, talking on the phone, and fetching objects. It also includes focused driving scenarios where subjects are attentive to side mirrors and the road. The balanced composition of 50\% distracted driving and 50\% focused driving samples across all three scenarios enhances the robustness of the distraction detection model. This comprehensive approach ensures that our model is robustly trained and capable of detecting distracted driving with high accuracy.

\subsection{Implementation Configuration}

The experiments were conducted on an Nvidia RTX 2080 Ti GPU with 64 GB of RAM, utilizing the PyTorch framework. During the training process for the simulated Driver Monitoring Dataset (DMD), the Adam optimizer was employed with an initial learning rate of 0.1. This learning rate was reduced by a factor of 0.1 every four epochs to ensure optimal performance. The training was stopped at 30 epochs.

\section{Results}


The table \ref{table1} presents a comparison of different models used for detection tasks, highlighting their respective modalities, accuracies, and parameter counts. The models compared include MobileNetv1+LSTM, Video Swin-Transformer, 3D-CNN, Submanifold-ResNET, and the Proposed Spike-DD. All models except Submanifold -ResNET and Proposed Spike-DD use RGB modality. The MobileNetv1+LSTM achieves an accuracy of 97.30\% with 5.51 million parameters, whereas the Video Swin-Transformer slightly outperforms it with 97.50\% accuracy but requires significantly more parameters (28.00 million). The 3D-CNN, with 2.93 million parameters, achieves an accuracy of 97.20\%. For event-based models, Submanifold-ResNET achieves an accuracy of 80.05\% with 0.311 million parameters, while the Proposed Spike-DD stands out by achieving a high accuracy of 94.40\% with a similar number of parameters (0.301 million), demonstrating a highly efficient performance relative to its parameter count.

In contrast to the visible modality models that rely on RGB images, which are typically evaluated using a multi-class classification approach with 4-11 distraction labels on the DMD dataset, the Proposed Spike-DD is a binary classifier designed specifically for detecting driver distraction. The significant performance of the Proposed Spike-DD highlights the advantages of using SNNs for this task. SNNs are uniquely suited for capturing and processing temporal information due to their event-driven nature, which aligns well with the dynamic nature of driver distraction scenarios. Unlike conventional methods that process entire frames of video data at once, the SNN's approach of handling discrete spikes over time allows it to exploit the temporal structure of the data. This temporal processing capability is crucial for detecting subtle changes in driver behavior over short time windows, which are characteristic of distraction events.

The design of the Proposed Spike-DD leverages this temporal information by translating events into a spiking format that preserves the timing of inputs, rather than processing each frame independently. This temporal coding allows the network to react to and differentiate between transient distractions more effectively. The inherent efficiency of SNNs also contributes to the model’s performance. By utilizing fewer parameters and an asynchronous processing approach, the Proposed Spike-DD not only achieves high accuracy but does so with a lower computational cost compared to traditional models. This efficiency is particularly important in real-time applications, where the ability to process data quickly and with minimal resource consumption can significantly impact the deployment of driver distraction detection systems.

Overall, the results in Table \ref{table1} underscore that the Proposed Spike-DD not only matches but exceeds the performance of state-of-the-art models in terms of accuracy while operating with fewer parameters. This demonstrates the potential of SNNs for real-time driver distraction detection and highlights how temporal information processing in short time windows can enhance the effectiveness of such models. The proposed network exemplifies how SNNs can leverage temporal dynamics to achieve both high accuracy and efficiency, setting a new benchmark for future research in this area.

\begin{table}[h!]
\centering
\begin{tabular}{|>{\columncolor[HTML]{003366}}l |>{\columncolor[HTML]{003366}}c |>{\columncolor[HTML]{003366}}c |>{\columncolor[HTML]{003366}}l |}
\hline
\textcolor{white}{\textbf{Model}} & \textcolor{white}{\textbf{Modality}} & \textcolor{white}{\textbf{Accuracy}} & \textcolor{white}{\textbf{Parameters (millions)}} \\ \hline
\rowcolor[HTML]{EAD1DC} 
MobileNetv1+LSTM \cite{canas2021detection}    & RGB-frame   & 97.30\%   & 05.51 M   \\ \hline
\rowcolor[HTML]{EAD1DC} 
Video Swin-Transformer \cite{lakhani2022applying} & RGB-frame   & 97.50\%   & 28.00 M  \\ \hline
\rowcolor[HTML]{EAD1DC} 
3D-CNN     \cite{canas2021detection}           & RGB-frame   & 97.20\%   & 02.93 M   \\ \hline
\rowcolor[HTML]{D9EAD3} 
Submanifold-ResNET  \cite{shariff2023neuromorphic}  & Event-Histograms & 80.05\%   & 0.311 M   \\ \hline
\rowcolor[HTML]{D9EAD3} 
Proposed Spike-DD     & Event-Spikes & 94.40\%   & 0.301 M  \\ \hline
\end{tabular}
\caption{The proposed model compared with state-of-the-art visible models and event camera-based models. Note: RGB models are multi-class, while event-based models are binary classifiers.}
\label{table1}
\end{table}

\subsection{Limitations}

While the Proposed Spike-DD model demonstrates substantial computational efficiency advantages, it is essential to recognize that Spiking Neural Networks (SNNs) are highly sensitive to hyperparameter tuning, which is crucial for achieving optimal performance in practical Driver Monitoring System (DMS) applications. The effectiveness of SNNs is influenced by several key hyperparameters, including the spike time window, neuron model parameters such as membrane resistance and threshold potential, learning rates, and the encoding scheme for event data. Each of these parameters (are mentioned in section 3.2) requires careful fine-tuning to balance temporal resolution, learning stability, and computational efficiency. Although our model shows promising results in terms of efficiency and performance, further research is needed to perform extensive testing and parameter optimization in real-world environments to fully validate the potential of SNNs for practical DMS deployments.

\section{Conclusion and Future Work}

This research presents a novel network methodology for the detection of driver distraction utilizing event-based data, specifically through the simulated DMD dataset (v2e). The method employs a spiking neural network, which is particularly effective in handling data with high temporal resolution, ensuring the swift and accurate detection of distractions. This network is characterized by a low parameter count, which minimizes computational requirements, making it ideal for real-time deployment. Additionally, the approach integrates a privacy-by-design framework to protect sensitive driver information throughout the detection process. The proposed Spiking-DD networks have outperformed existing models in both accuracy and efficiency due to their reduced parameter needs. Overall, the combination of event-based data with spiking neural networks represents a promising, efficient, and privacy-conscious strategy to enhance driver safety. Future work will involve evaluating the real-time capabilities of this approach and its performance on neural accelerators such as the Loihi-2 chip from Intel. Additionally, more extensive testing in hardware and varying environmental conditions (e.g., different lighting levels) is needed to fully validate the potential of SNNS.

\section*{Acknowledgement}
The authors express their gratitude to the Irish Research Council, FotoNation Ltd, and the University of Galway. This study is funded by the Irish Research Council under grant number EBPPG/2022/17.
\appendix

\bibliographystyle{apalike}

\bibliography{imvip}

\end{document}